\pdfoutput=1

\documentclass[11pt]{article}

\usepackage[review]{acl} 

\usepackage{times}
\usepackage{latexsym}

\usepackage[T1]{fontenc}

\usepackage[utf8]{inputenc}

\usepackage{microtype}

\usepackage{inconsolata}

\usepackage{graphicx}

\usepackage{kotex}
\usepackage{booktabs}
\usepackage{multirow}
\usepackage{enumitem}
\setitemize{noitemsep,topsep=2pt,parsep=0pt,partopsep=0pt}
\usepackage{amsmath}
\usepackage{amssymb}
\usepackage{float}
\usepackage[table]{xcolor}
\usepackage{colortbl}
\usepackage{threeparttable}
\usepackage{stfloats}
\usepackage{tablefootnote}
\usepackage{tocloft} 
\usepackage{hyperref}

\title{ELO: Efficient Layer-Specific Optimization for Continual Pretraining of Multilingual LLMs}
\newcommand\CoauthorMark{\footnotemark[\value{footnote}]}

\author{
    Hangyeol Yoo$^1$\thanks{~Equal Contribution}\hspace{2.5mm}
    ChangSu Choi$^{1,2}$\protect\CoauthorMark\thanks{~Work done during an internship at LG CNS}\hspace{2.5mm}
    \bf Minjun Kim$^3$\hspace{2.5mm}
    Seohyun Song$^1$\\
    \bf SeungWoo Song$^1$\hspace{2.5mm}
    Inho Won$^3$ \hspace{2.5mm}
    Jongyoul Park$^1$\hspace{2.5mm}
    Cheoneum Park$^4$\hspace{2.5mm}
    KyungTae Lim$^3$\thanks{~Corresponding Author} \\
    $^1$Seoul National University of Science and Technology \hspace{0.5mm}
    $^2$LG CNS \\
    $^3$Korea Advanced Institute of Science and Technology \hspace{0.5mm}
    $^4$Hanbat National University\\
    \texttt{\{hgyoo, choics2623\}@seoultech.ac.kr}, \texttt{ktlim@kaist.ac.kr}
}

\begin{document}

\maketitle 

\addtocontents{toc}{\protect\setcounter{tocdepth}{0}}

\maketitle
\begin{abstract}
We propose an efficient layer-specific optimization (ELO) method designed to enhance continual pretraining (CP) for specific languages in multilingual large language models (MLLMs). This approach addresses the common challenges of high computational cost and degradation of source language performance associated with traditional CP. The ELO method consists of two main stages: (1) ELO Pretraining, where a small subset of specific layers, identified in our experiments as the critically important first and last layers, are detached from the original MLLM and trained with the target language. This significantly reduces not only the number of trainable parameters but also the total parameters computed during the forward pass, minimizing GPU memory consumption and accelerating the training process. (2) Layer Alignment, where the newly trained layers are reintegrated into the original model, followed by a brief full fine-tuning step on a small dataset to align the parameters. Experimental results demonstrate that the ELO method achieves a training speedup of up to 6.46 times compared to existing methods, while improving target language performance by up to 6.2\% on qualitative benchmarks and effectively preserving source language (English) capabilities.
\end{abstract}
\section{Introduction}
\label{sec:Introduction}
Recent studies have focused on enhancing multilingual large language models (MLLMs) for specific languages~\cite{zhao2023survey}. Notably, studies like Chinese-Llama~\cite{chinese-llama} and EEVE~\cite{eeve} have demonstrated improved performance by continual pretraining (CP) of MLLMs for target languages. However, these models encounter two major challenges. First, enhancing performance on a target language often significantly degrades performance on the primary language, English~\cite{choi-etal-2024-optimizing-language}. Second, enhancing performance in the target language through CP demands significant time and resources, posing challenges for small-scale researchers~\cite{naveed2024comprehensive}. To address these issues, lightweight training techniques such as Low-Rank Adaptation (LoRA) have been introduced to enhance model performance by modifying only a portion of the model~\cite{lora}. However, even when using LoRA, the time savings compared with full fine-tuning (FFT) are minimal. This is because while it significantly reduces the number of trainable parameters, the forward pass requires computation through both the original model weights and the additional LoRA parameters.


This computational overhead during the forward pass led us to a new hypothesis. Instead of merely limiting the trainable parameters within the full model (like LoRA), what if we could also reduce the computed parameters during CP by training a much smaller, separate model?

This line of inquiry led to our core concept: detaching a small subset of MLLM layers to be trained independently. Following this, we propose an efficient layer-specific optimization (ELO) method that focuses solely on this detached portion of layers for enhancing specific languages. The proposed method comprises two phases: ELO pretraining and layer alignment. First, ELO pretraining involves this detachment and CP process to imbue specific linguistic knowledge. Layer alignment is the phase where the newly acquired knowledge from ELO pretraining is transferred into the original MLLM.

This approach significantly reduces the number of model parameters during CP, thereby minimizing time and resource costs. Experimental results indicate that the training speed of the proposed method was 6.46 times faster. Qualitative evaluations performed similar or up to 6.2\% superior results. The contributions of this study can be summarized as follows:
\begin{itemize}
\item We propose an efficient CP method, ELO, for MLLMs, enriching the availability of specific languages.
\item Through comprehensive analysis, we have demonstrated the real-world effectiveness of the approach employed in our method.
\end{itemize}

\section{Related Work}
\label{sec:Related Work}

\paragraph{Efficient Fine-Tuning.}
Parameter-efficient fine-tuning (PEFT) methods are gaining prominence as language models continue to grow in size. These methods efficiently customize pretrained models for specific languages or tasks~\cite{bai2024efficiency}. Among these, LoRA~\cite{lora+,relora,qlora} is a notable lightweight training method that achieves performance comparable to that of FFT by training a subset of parameters.
However, these methods offer minimal training speedup over FFT. This is because while they reduce the number of trainable parameters, the computational cost remains high, as the forward pass must still be computed through all original model weights and the additional adapter parameters~\cite{lora}.

\paragraph{Selective Layer Tuning.}
As the number of layers in LLM increases, research on layer-selective tuning, based on the distinct roles each layer performs, has been proposed.
\citet{lad2024remarkablerobustnessllmsstages} demonstrated that not all layers serve the same function; the middle layers are responsible for understanding context and sentence structure, whereas the initial and final layers focus on integrating information. In a similar vein, EEVE~\cite{eeve} proposed a method for training only specific layers in the target language to improve performance in target language.
While selective, this approach (as utilized in EEVE) still operates within the full model architecture. Consequently, it suffers from the same computational overhead as LoRA: the forward pass must still be computed across all model parameters, even if only a subset of layers is being updated~\cite{eeve}.

\section{Efficient Layer-Specific Optimization}
\label{sec:Efficient_Layer-specific_Optimization}

As established in Section~\ref{sec:Related Work}, conventional PEFT methods like LoRA and selective layer training suffer from a significant computational bottleneck. Although they reduce the number of trainable parameters, they still require the forward pass to be computed across the entire model architecture. This results in minimal training speedup over FFT.

To overcome this fundamental limitation, we propose Efficient Layer-Specific Optimization (ELO). The core idea of ELO is to detach a small subset of specific layers from the original model before pretraining. This action creates a much smaller, independent model for the CP phase. This  layer detachment approach directly solves the overhead problem by drastically reducing not only the trainable parameters but also the total parameters computed during the forward pass. The ELO method comprises two main stages: (1) ELO pretraining and (2) layer alignment.

\begin{figure} 
\scriptsize
  \includegraphics[width=\linewidth]{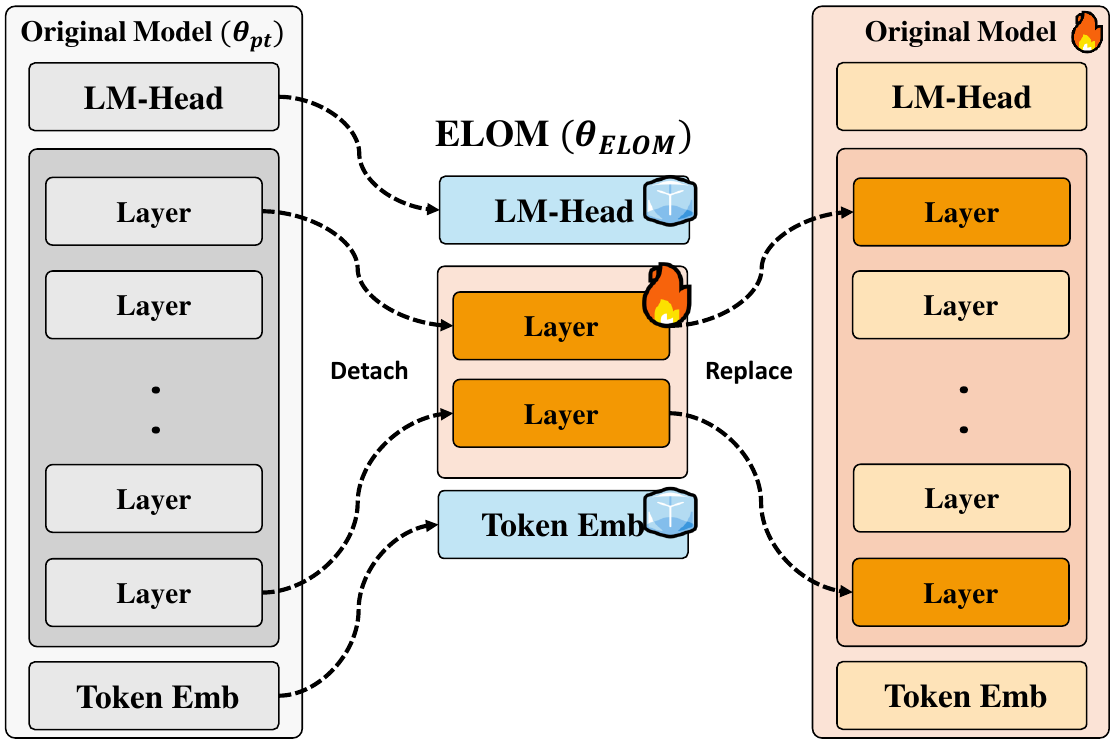}
  \caption{Description of the proposed ELO training process}
  \label{fig:elo_arch}
\end{figure}
\subsection{ELO Pretraining}

The initial stage involves detaching specific layers from the original model for pretraining, as shown in Figure~\ref{fig:elo_arch}. The language model comprises \( n \) decoder layers \( \mathbf{L} = \{\ell_1, \ell_2, \ldots, \ell_n\} \), the token embedding layer \( \ell_e \), and the head layer \( \ell_h \). We define the set of specific layers that comprise the ELO model as \( \lambda \subset \mathbf{L} \), where \( \theta^e \) and \( \theta^h \) represent parameters for \( \ell_e \) and \( \ell_h \), respectively, and \( \theta^\lambda \) represents parameters for \( \lambda \). We selected \( \lambda \) to encompass the first and last decoder layers, i.e., \(\lambda = \{\ell_1, \ell_n\}\). The pretraining process can be expressed as follows:
\begin{align}
    \theta_{\text{ELOM}} &= \{\theta^{e}, \theta^{h}, \theta^{\lambda}\}\\
    \mathcal{L}_{\text{PT}} &= -\sum_{i=1}^{|D_{\text{pt}}|} \sum_{j=1}^{|x_i|} \log{P(x_j | x_{<j}; \theta_{\text{ELOM}})}
    \label{eqa:pretraining}
\end{align}

The ELO model was trained using each sample from the pretraining dataset \( D_{pt} \), with the English-\{Target Language\} ratio set to 1:9 according to Equation~\ref{eqa:pretraining}. In this context, \( \mathcal{L}_{PT}\) represents the causal language-modeling loss function, with \( \theta_{0} \) denoting the parameters of the original model. Only the parameters \( \theta^{\lambda} \) of the ELO model are trained to infuse knowledge of the target language.

\subsection{Layer Replacement and Alignment}
\label{sec:extract_elo_vector&Layer_Aligning}
The second stage transfers knowledge learned during the first stage back to the original model $(\theta_0)$. We replace \( \theta^{\lambda} \) in the original model \( \theta_0 \) with $\theta_{\text{ELOM}}$. By replacing these two layers, it is possible to inject knowledge of the target language into specific layers while preserving the finely tuned token embedding and head layers, as well as existing layers rich in English knowledge. However, because this method modifies only the parameters of specific layers in the original model, aligning these layers requires further pretraining using a small dataset. Accordingly, we introduce a layer alignment step after replacement, wherein FFT is applied to the entire model using an additional 1GB of training data. 



\subsection{Bilingual Instruction Tuning}
\label{sec:Instruction_Tuning}
Because the model that has undergone the layer alignment process is a pretrained model, it exhibits limited capability in following user instructions. 
To efficiently improve instruction-following performance in the target language with less data, we first adopt the chat vector method~\cite{huang2024chat}. 

This method extracts knowledge by calculating the deviation (\(\theta_{\text{chat vector}} = \theta_{\text{Inst}} - \theta_{\text{PT}}\)) between a pretrained model (\(\theta_{\text{PT}}\)) and an instruction-tuned model (\(\theta_{\text{Inst}}\)). 
The extracted \(\theta_{\text{chat vector}}\) is then integrated into our layer-aligned model to efficiently transfer the instruction-following capabilities. 
After integrating the chat vector, we then conducted supervised fine-tuning (SFT).

For SFT, we utilized 31K instruction data extracted in a 1:1 ratio in both the target language and English from the ShareGPT-style dataset~\cite{devine2024tagengo}, which contains dialogue records from language models such as GPT4~\cite{openai2023gpt4}. 
Details of the dataset can be found in Appendix~\ref{sec:appendix-IT}.
\section{Experiments}
\label{sec:Experiments}
Our experiments are designed to empirically validate the main claims of ELO. We aim to assess whether our layer detachment strategy successfully overcomes the forward-pass bottleneck and leads to significant training speedups compared to FFT and LoRA. We also evaluate whether ELO effectively enhances performance in the target languages (Korean and Japanese) compared to both the base model and traditional FFT, and whether it maintains strong performance in the source language (English) without the significant degradation often seen in CP. To answer these questions, we first introduce the evaluation benchmarks, the models used, and then present our main results. We follow this with an in-depth ablation study to justify our specific design choices, such as layer selection and the alignment process.

\begin{table*}[t]
\centering
\tiny
\resizebox{\textwidth}{!}{
    \begin{tabular}{lrrrrrrcccccc}
        \toprule
        \multicolumn{1}{c}{Model} & \multicolumn{2}{c}{CP / ELO pretraining}& \multicolumn{2}{c}{Layer alignment} & & \multicolumn{2}{c}{Quantitative Eval.} & \multicolumn{2}{c}{Qualitative Eval. (Out of 10)} \\
        \hline 
        \rowcolor{gray!20}\multicolumn{10}{c}{\textbf{Korean}} \vspace{0.05cm}\\
        & Data size~(Tokens) & Time & Data size~(Tokens) & Time & Total time & MMLU\textbf{(en)} & KoBEST\textbf{(ko)} & MT-Bench\textbf{(en)} & LogicKor\textbf{(ko)}\\
        \midrule
        \href{https://huggingface.co/meta-llama/Meta-Llama-3.1-8B-Instruct}{\texttt{Llama3.1-8B-Instruct}}           & -     & -& -       & -       & -         & \textbf{68.07} & 56.02& \textbf{6.96}   & 6.03\\
        \href{https://huggingface.co/meta-llama/Meta-Llama-3.1-8B}{\texttt{Llama3.1-FFT}}                   & 10~GB~(2.8~B) & 19.8~h & -  & -  & 19.8~h    & 67.51          & \textbf{66.08}   & 6.70& 7.31\\
        \href{https://huggingface.co/meta-llama/Meta-Llama-3.1-8B}{\texttt{Llama3.1-ELO}}                   & 9~GB~(2.5~B)  & 1.5~h& 1~GB~(0.3~B)   & 2.0~h   & 3.5~h     & 66.69          & 60.81            & 6.79& \textbf{7.76}      \\
        \midrule
        \href{https://huggingface.co/mistralai/Mistral-7B-Instruct-v0.3}{\texttt{Mistral-7B-Instruct-v0.3}}       & -     & -& -       & -       & -         & \textbf{59.69} & 49.01            & 6.72& 4.49\\
        \href{https://huggingface.co/mistralai/Mistral-7B-v0.3}{\texttt{Mistral-FFT}}                    & 10~GB~(4.7~B) & 31.0~h& -  & -  & 31.0~h    & 57.47          & \textbf{61.68}   & 6.61&  6.50    \\
        \href{https://huggingface.co/mistralai/Mistral-7B-v0.3}{\texttt{Mistral-ELO}}                    & 9~GB~(4.2~B)  & 1.7~h& 1~GB~(0.6~B)  & 3.1~h   & 4.8~h     & 58.90          & 60.56& \textbf{6.97}   & \textbf{6.59}\\
        \midrule
        \href{https://huggingface.co/Qwen/Qwen2-7B-Instruct}{\texttt{Qwen2-7B-Instruct}}              & -     & -       & -& -       & -         & 69.89          & 60.17            & \textbf{7.67}   & 6.90               \\
        \href{https://huggingface.co/Qwen/Qwen2-7B}{\texttt{Qwen2-FFT}}                      & 10~GB~(3.1~B) & 20.5~h& -  & -  & 20.5~h    & \textbf{70.23} & \textbf{72.37}   & 7.11            & 6.95               \\
        \href{https://huggingface.co/Qwen/Qwen2-7B}{\texttt{Qwen2-ELO}}                      & 9~GB~(2.8~B)  & 1.8~h& 1~GB~(0.3~B)   & 2.1~h   & 3.9~h     & 70.11          & 71.57            & 7.25            & \textbf{7.22}      \\

\rowcolor{gray!20}\multicolumn{10}{c}{\textbf{Japanese}} \vspace{0.05cm}\\
        & Data size~(Tokens) & Time & Data size~(Tokens) & Time & Total time & MMLU\textbf{(en)} & MARC-\textbf{ja} & MT-Bench\textbf{(en)} & MT-Bench\textbf{(ja)}\\
        \midrule
        \href{https://huggingface.co/meta-llama/Meta-Llama-3.1-8B-Instruct}{\texttt{Llama3.1-8B-Instruct}}           & -     & -& -       & -       & -         & \textbf{68.07}  & \textbf{96.36}  & 6.96          & 4.85            \\
        \href{https://huggingface.co/meta-llama/Meta-Llama-3.1-8B}{\texttt{Llama3.1-FFT}}                   & 10~GB~(2.7~B) & 19.4~h & -  & -  & 19.4~h    & 67.50           & 96.25           & \textbf{6.99} & 5.38            \\
        \href{https://huggingface.co/meta-llama/Meta-Llama-3.1-8B}{\texttt{Llama3.1-ELO}}                   & 9~GB~(2.4~B)  & 1.4~h & 1~GB~(0.3~B)   & 2.0~h   & 3.4~h     & 67.51           & 95.35           & 6.90          & \textbf{5.58}   \\
        \midrule
        \href{https://huggingface.co/mistralai/Mistral-7B-Instruct-v0.3}{\texttt{Mistral-7B-Instruct-v0.3}}       & -     & -& -       & -       & -         & \textbf{59.69}  & 83.43           & \textbf{6.72 }& 4.36             \\
        \href{https://huggingface.co/mistralai/Mistral-7B-v0.3}{\texttt{Mistral-FFT}}                    & 10~GB~(4.1~B) & 29.7~h & -  & -  & 29.7~h    & 54.86           & 80.05           & 6.26          & \textbf{5.68}    \\
        \href{https://huggingface.co/mistralai/Mistral-7B-v0.3}{\texttt{Mistral-ELO}}                    & 9~GB~(3.7~B)  & 1.7~h & 1~GB~(0.4~B)   & 3.0~h   & 4.7~h     & 55.19           & \textbf{89.53}  & 6.38          & \textbf{5.68}    \\
        
        \bottomrule
    \end{tabular}
}
\caption{Performance and training time comparison of the proposed ELO method and FFT for three languages.}
\label{tab:korean_japanese_eval}
\end{table*}

\subsection{Evaluation Benchmarks}
\label{sec:benchmarks}

We conducted experiments to evaluate the effectiveness of ELO using Korean and Japanese as target languages, chosen for their distinct differences from English. Our evaluation of the LLMs was divided into quantitative and qualitative assessments~\cite{zhou2023lima,choi-etal-2024-optimizing-language}. Quantitative evaluation involves scoring based on numerical metrics (e.g., accuracy, F1-score), while qualitative evaluation assesses long-form generative answers using an LLM-as-a-judge (e.g., GPT-4).

For English, we used \textbf{MMLU}\cite{hendrycks2020measuring} for quantitative evaluation, a benchmark that evaluates knowledge across 57 topics, measuring accuracy. For qualitative evaluation, we used \textbf{MT-Bench}~\cite{NEURIPS2023_91f18a12}, a set of 80 challenging multi-turn open-ended questions evaluated by a GPT-4 judge on a 10-point scale.

For Korean, the quantitative benchmark was \textbf{KoBEST}~\cite{jang-etal-2022-KoBEST}, a suite of 5 NLU tasks requiring advanced Korean knowledge, evaluated using F1-score. The qualitative benchmark was \textbf{LogicKor}~\cite{logickor}, a multi-turn dataset measuring reasoning ability across six domains (e.g., reasoning, mathematics, coding) with 42 prompts, also judged by GPT-4 on a 10-point scale.

For Japanese, we employed \textbf{MARC-ja}~\cite{marc_ja} for quantitative assessment, a text classification task based on the Multilingual Amazon Reviews Corpus, using accuracy\_norm as the metric. For qualitative assessment, we used \textbf{MT-Bench(ja)}~\cite{Japanese-MT-Bench}, a Japanese translation of MT-Bench, which similarly uses a GPT-4 judge and a 10-point scale.

\subsection{Model Description}
We compared the proposed ELO method with conventional FFT in terms of efficiency using the following open-source LLMs: Llama 3.1-8B~\cite{llama3.1}, Mistral-7B-v0.3~\cite{mistral}, and Qwen2-7B ~\cite{qwen2}. The model names in Table~\ref{tab:korean_japanese_eval} refer to models trained using the following methods:

\paragraph{\texttt{\{base\_model\}-Instruct}} The official instruct-tuned models are released by each company. 

\paragraph{\texttt{\{base\_model\}-FFT}} This model refers to one that was first fine-tuned on the \texttt{base\_model} using the FFT method, followed by instruction tuning, as outlined in Section~\ref{sec:Instruction_Tuning}. For example, the \texttt{Llama3.1-FFT} model in Table~\ref{tab:korean_japanese_eval} was trained with 10GB of CP data, followed by instruction tuning with 31K data.

\paragraph{\texttt{\{base\_model\}-ELO}} This refers to a model that applied the ELO method proposed in Section~\ref{sec:Efficient_Layer-specific_Optimization}.
\subsection{Experimental Results}
\label{sec:Experiment-Results}

\paragraph{Overall.} The results presented in Table~\ref{tab:korean_japanese_eval} indicate that both the FFT and ELO configurations significantly outperformed the \texttt{\{base\_model\}-Instruct} models in the qualitative evaluation. Specifically, the ELO method achieved a 22.2\% improvement in LogicKor performance compared with \texttt{Llama3.1-8B-Instruct}. However, in the quantitative evaluations, performance varied considerably across languages and base models. These findings indicate that the proposed pretraining and bilingual instruction tuning methods significantly enhance performance on target languages.

\paragraph{Qualitative Evaluation Effect of ELO.} 
As shown in the Qualitative Evaluation column of Table~\ref{tab:korean_japanese_eval}, the models trained with the ELO consistently outperformed those trained with FFT in the qualitative assessments for Korean, and Japanese. Notably, the ELO models achieved higher scores in LogicKor, with a 0.45p improvement for Llama3.1 and a 0.27p improvement for Qwen2, than their FFT counterparts. Also, ELO has demonstrated substantial efficiency, reducing the training time by an average of 5.88-fold compared with that of FFT.

\paragraph{Quantitative Evaluation Effect of ELO.}
In the quantitative evaluations, performance varied with respect to the source language (English) and target languages. For the English MMLU evaluation, the base (-Instruct) models generally achieved the highest performance. However, the average performance difference compared with ELO was only 2.42\%, suggesting that the impact was minimal. This is likely because both ELO and FFT were more focused on target languages using a 1:9 ratio in CP. Supporting this, the Korean quantitative evaluation (KoBEST) results show that the ELO models consistently outperformed the base (-Instruct) models by margins ranging from 8.55\% to 23.57\%.




\paragraph{Resource Efficiency of ELO.}
ELO has demonstrated substantial efficiency, reducing the training time by an average of 5.88-fold compared with FFT. The strength of ELO lies in its ability to achieve comparable or superior qualitative performance to that of FFT while using fewer computational resources. When trained on 10GB of PT data, ELO accelerates training by 5.26 to 6.46 times compared to FFT. For example, as shown in Table 1, the ELO-enabled model outperformed the Llama3.1-FFT model by 6.2\% on LogicKor while achieving a 5.66-fold speedup.

Furthermore, Figure 3 shows that ELO significantly outperforms LoRA in training speed, empirically validating our hypothesis from Sections 1 and 2. While Figure 3 confirms that LoRA provides minimal time savings over FFT, ELO is 5.29 times faster than LoRA when trained with 50GB of data. This efficiency gap widens as the data size increases; with 200GB of data, ELO is 10.72 times faster than LoRA. These results demonstrate that ELO's layer detachment strategy successfully overcomes the forward-pass computational bottleneck that limits LoRA.

\section{Ablation Study}
\label{sec:Analysis}
In Section~\ref{sec:Experiments}, we demonstrated that ELO achieves superior efficiency and performance compared to existing methods. However, critical questions regarding the optimal configuration and the underlying mechanisms of ELO remain. In this section, we conduct an in-depth analysis using the Korean qualitative benchmark, LogicKor, to address these inquiries. We first investigate whether the performance gain scales with the amount of pretraining data or if the limited capacity of the detached layers poses a bottleneck. We then examine if the improvements are consistent across different model sizes, such as 70B parameters, and disentangle the contribution of bilingual instruction tuning from the ELO pretraining itself. Furthermore, we provide an empirical justification for our selection of the first and last layers and analyze the sensitivity of performance to this choice. Finally, we verify the necessity of the layer alignment phase and determine the optimal amount of data required for this step.

\paragraph{ELO with More Pretraining.} 
We now raise the question of whether increasing the volume of pretraining data diminishes learning effectiveness owing to the limited capacity of the layers to accommodate information. After examining the results in Table~\ref{tab:LogicKor_inner}, we evaluated the performance of the \texttt{Llama3-ELO} model by pretraining it with volumes of data ranging from 10 to 200GB. Based on these findings, we observed that as the volume of pretraining data increased, there were substantial improvements in performance.

\begin{table}[t]
\tiny
\centering
\begin{tabular}{lrrccc}
\toprule
\textbf{Model}      & \textbf{Param} & \textbf{Data} & \textbf{Single} & \textbf{Multi} &\textbf{Total} \\
\midrule 
\texttt{Llama3-8B-Instruct}      & 8~B & - & 2.09  & 2.54 & 2.32 \\
\texttt{Llama3-8B-Instruct-SFT}      & 8~B & -      & 6.26 & 5.45 & 5.86 \\
\texttt{Llama3-FFT}      & 8~B & 10~GB  & 6.14 & 6.21 & 6.18 \\
\texttt{Llama3-ELO}         & 8~B & 10~GB  & 6.40 & 5.95 & 6.18 \\
\texttt{Llama3-ELO}         & 8~B & 50~GB  & 6.40 & 6.36 & 6.38 \\
\texttt{Llama3-ELO}         & 8~B & 200~GB & \textbf{6.95} & \textbf{7.00}& \textbf{6.97} \\
\midrule
\texttt{Llama3-70B-Instruct}      & 70~B & -     & 2.62 & 3.00 & 2.76 \\
\texttt{Llama3-ELO}         & 70~B & 50~GB & 7.52 & 7.24 & 7.38 \\
\texttt{Llama3.1-70B-Instruct}      & 70~B & -     & 7.66 & 7.90 & 7.78 \\
\texttt{Llama3.1-ELO}            & 70~B & 50~GB  & \textbf{8.79} & \textbf{8.52} & \textbf{8.65} \\
\bottomrule
\end{tabular}
\centering 
\caption{Internal evaluation results using LogicKor.}
\label{tab:LogicKor_inner}
\end{table}


\paragraph{ELO with Bigger Size Model.} 
Another question regarding the ELO method is whether similar performance improvements can be observed in larger models. The performance results for the 70B model are shown in Table~\ref{tab:LogicKor_inner}. A comparison between \texttt{Llama3-70B-Instruct} and \texttt{Llama3-ELO}, both based on the 70B model, demonstrated significant performance improvements with the ELO model. However, since Llama3.1 showed substantial improvements in Korean language performance compared to version 3.0, additional experiments were needed to compare \texttt{Llama3.1-70B-Instruct} and \texttt{Llama3.1-ELO}. These experiments also revealed a notable performance increase of 10\% with ELO.

\paragraph{Impact of Bilingual Instruction Tuning.} 
In Table~\ref{tab:LogicKor_inner}, \texttt{Llama3-8B-Instruct-SFT} refers to the model fine-tuned on \texttt{Llama3-8B-Instruct} using the instruction data outlined in Section~\ref{sec:Instruction_Tuning}. This significant performance improvement highlights the impact of instruction data. Moreover, the performance gap between the ELO model, which uses relatively small amounts of PT data, and \texttt{Llama3-8B-Instruct-SFT} was minimal. This suggests that increasing the volume of PT data is crucial to fully leveraging the benefits of ELO.

\paragraph{Why were the first and last layers selected?}
\label{sec:appendix-ELO-layer}
\begin{table}[htbp]
\scriptsize
\begin{tabular}{lrrcc}
\toprule
\textbf{Model}      & \textbf{MT-Bench (en)} & \textbf{LogicKor (ko)}  \\
\midrule 
\texttt{Llama3.1-ELO(1,32)}         & 6.96 & 7.76 \\
\texttt{Llama3.1-ELO(1,16,32)}      & 7.11 & 7.69 \\
\texttt{Llama3.1-ELO(1,16)}         & 7.03 & 5.79 \\
\texttt{Llama3.1-ELO(8,24)}         & 7.19 & 5.00 \\
\texttt{Llama3.1-ELO(16,17)}        & 7.00 & 5.47 \\
\bottomrule
\end{tabular}
\centering 
\captionsetup{justification=centering}
\caption{Impact of layer selection on ELO.}
\label{tab:Impact_of_layer_selection_on_ELO}
\end{table}

Our experiments revealed that applying ELO to the first and last layers yields the best performance. As shown in Table~\ref{tab:Impact_of_layer_selection_on_ELO}, the proposed \( \lambda = \{\ell_1, \ell_{n}\} \) configuration, specifically \texttt{Llama3.1-ELO (1,32)}, significantly improved the LogicKor score to 7.76. This configuration, along with \texttt{Llama3.1-ELO (1,16,32)}, outperformed all others, indicating that the first and last layers are the most critical. This aligns with \citet{lad2024remarkablerobustnessllmsstages}'s findings, which highlight the importance of these layers in synthesizing and aggregating information.

In contrast, other configurations were less effective. Training intermediate layers, such as \( \lambda = \{\ell_8, \ell_{24}\} \) (\texttt{Llama3.1-ELO(8,24)}), resulted in a notably poor score of 5.0. This suggests that different layers vary in importance when incorporating new knowledge. Furthermore, using only the 1st and 16th layers (\texttt{Llama3.1-ELO(1,16)}) led to minimal improvements, suggesting that the first layer alone struggles to maintain consistent knowledge flow.

An interesting observation is that, regardless of which layers were trained with ELO, the MT-Bench (English) scores remained stable. This likely reflects the fact that the layers not involved in ELO training (e.g., 30 layers in the (1,32) configuration) retained their English knowledge, preserving performance. However, when ELO was trained exclusively on the target language without bilingual training, we observed a decline in performance.

\begin{figure}
\scriptsize
\includegraphics[width=\linewidth]{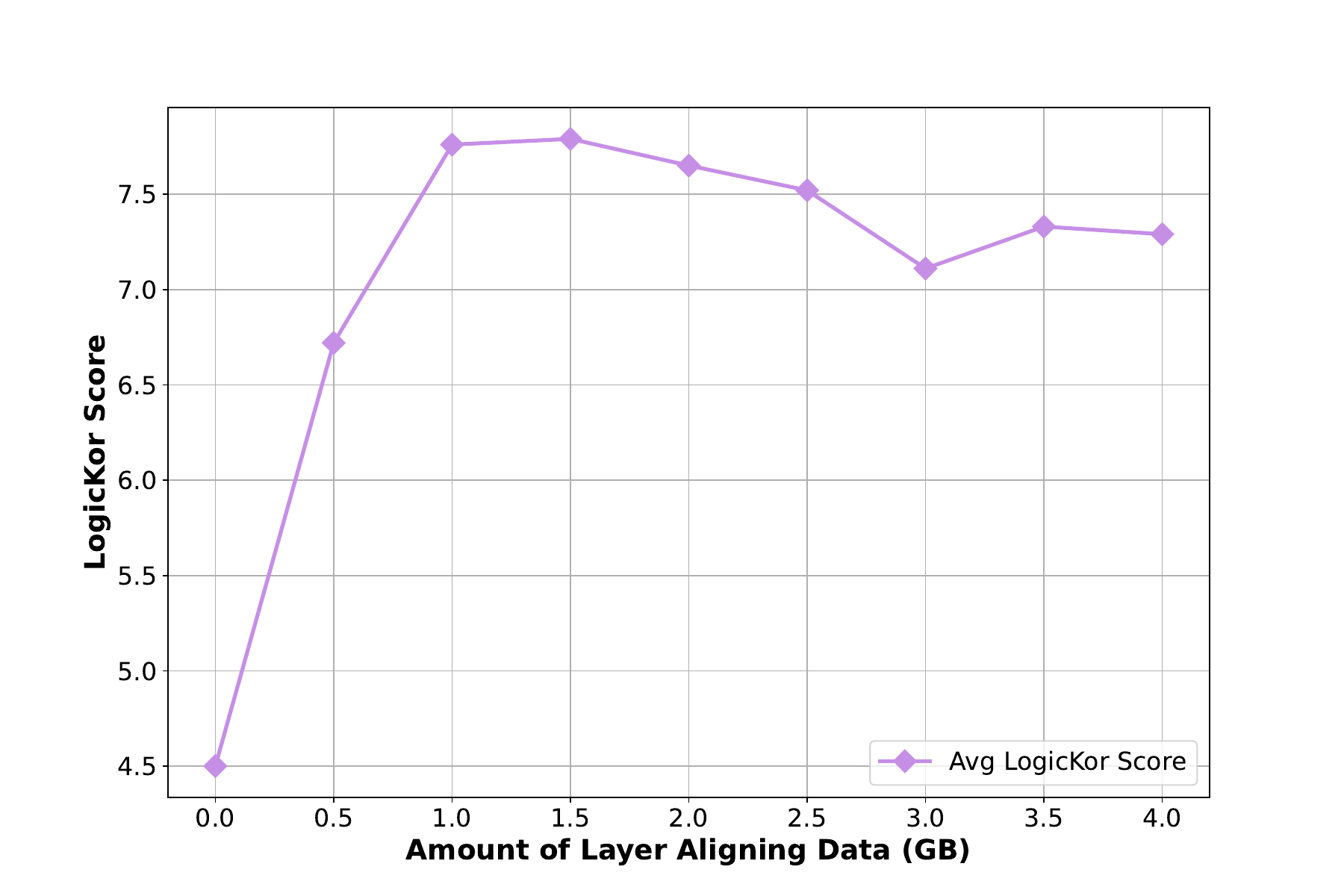}
\caption{Performance variation of LogicKor based on the amount of PT data for its layer aligning}
\label{fig:aligning_data_score}
\end{figure}

\paragraph{Effect of Layer Aligning}
\label{sec:appendix-Effect-of-Layer-Aligning}
To examine whether layer aligning is necessary and how much data is required for optimal performance, we progressively increased the amount of alignment data from 0GB to 4GB in 0.5GB increments using the \texttt{Llama-ELO} model described in Table~\ref{tab:korean_japanese_eval}. As shown in Figure~\ref{fig:aligning_data_score}, omitting layer aligning yielded the lowest LogicKor score (4.5), whereas even 1GB of data improved performance substantially to 7.76. The best result was obtained with 1.5GB (7.78), but further increases did not provide meaningful gains and in some cases slightly reduced performance. These results demonstrate that layer aligning is a crucial component of the ELO method and that only a small amount of bilingual data (approximately 1GB) is sufficient to achieve near-optimal performance. Consequently, we adopt 1GB as the default alignment size throughout this study to balance efficiency and effectiveness.

\begin{figure}
\scriptsize
\includegraphics[width=\linewidth]{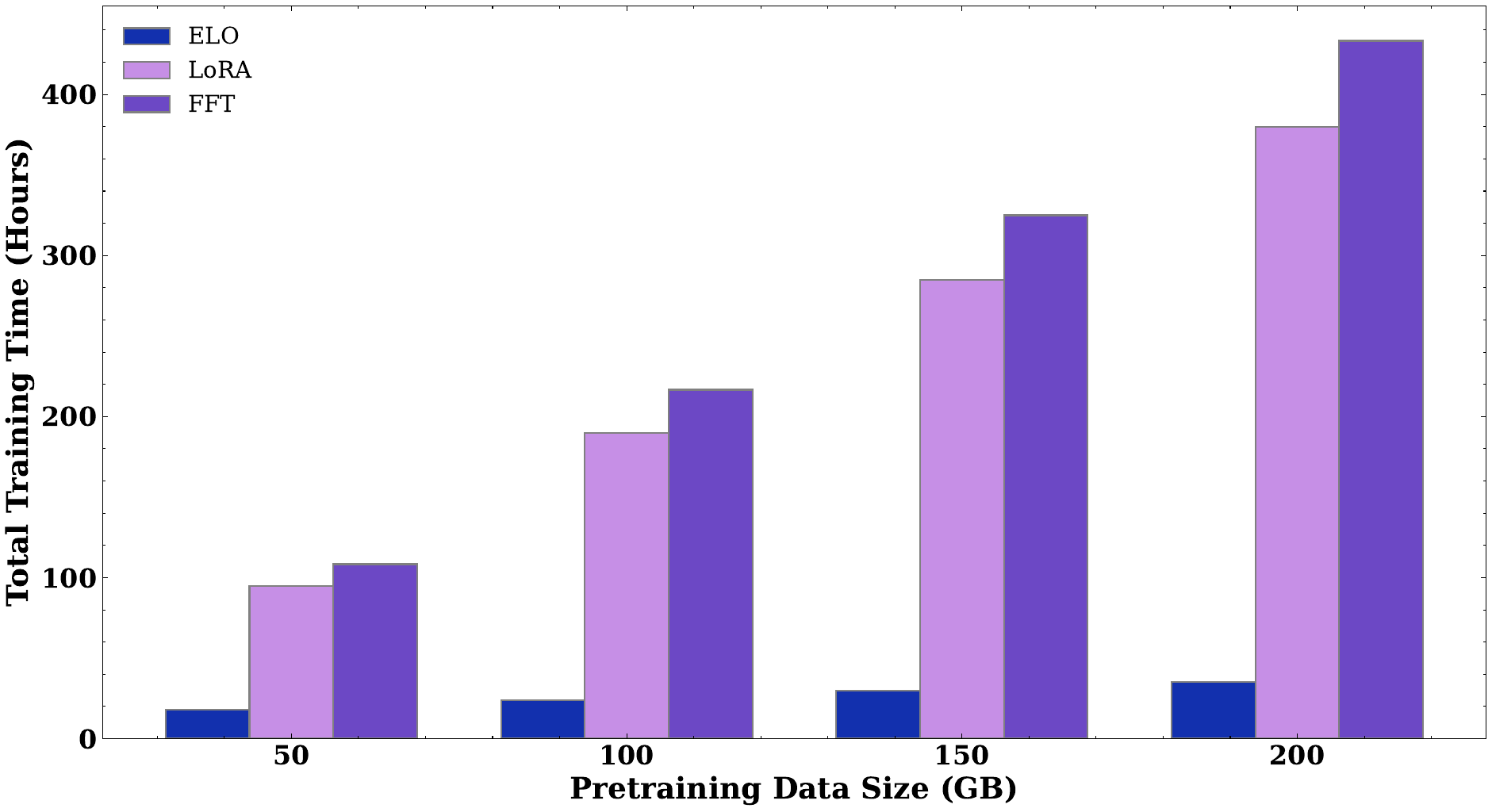}
\caption{Comparison of the training time across ELO, FFT, and LoRA training methods}
\label{fig:comparison_training_time}
\end{figure}

\paragraph{Comparison of Training Speed Based on Pretraining Data Size}
As mentioned in Section~\ref{sec:Related Work} and Section~\ref{sec:Efficient_Layer-specific_Optimization}, limiting the number of trainable parameters in a model does not significantly reduce training time unless the model's overall size is reduced. Additionally, as the amount of training data increases, larger models require substantially more training time compared to smaller models. Figure~\ref{fig:comparison_training_time} presents a comparison of the time taken to train \texttt{Llama3.1-8B} using ELO, LoRA, and FFT methods. When trained with 50GB of data, the ELO method is 5.29 times faster than LoRA and 6.04 times faster than FFT. However, when trained with 200GB of data, this difference increases to 10.72 times and 12.23 times. Therefore, the proposed method of enhancing specific languages through selective layer training becomes increasingly efficient as the amount of training data grows.

\section{Conclusion}
\label{sec:Conclusion}

In this paper, we proposed Efficient Layer-Specific Optimization (ELO) to address the computational bottleneck of continual pretraining (CP) in MLLMs. 
Existing PEFT methods like LoRA offer minimal training speedup because they must compute the forward pass across the entire model. ELO overcomes this via a layer detachment strategy By training a small subset of critical layers (the first and last) as a smaller, independent model, ELO drastically reduces the parameters computed during the CP phase. 
This approach minimizes GPU memory consumption during this pretraining phase and enables significant acceleration.

Our experimental results demonstrate that ELO achieves a training speedup of up to 6.46 times compared to FFT. It also yields superior qualitative performance in target languages by up to 6.2\%, while effectively preserving source language capabilities. 
This work establishes ELO as a highly efficient and effective alternative for multilingual adaptation.
\section{Limitations}
\label{sec:Limitations}

The ELO method minimizes GPU memory usage during the pretraining phase and accelerates the overall training process; however, it still has the following two limitations.

\paragraph{Even a minimal amount of FFT is required} 
While our experiments have shown that layer alignment with a minimal amount of data, such as 1GB, is sufficient, the layer alignment phase remains essential. Since this phase requires training all the parameters of the original model, it demands more GPU memory than ELO pretraining. Therefore, it does not reduce the peak GPU memory requirement in the overall training process.

\paragraph{Investigation of CP experiments with over 1TB of data} 
The performance of the FFT and ELO methods has not been verified when the data size exceeds 1TB. Unfortunately, it was impossible to conduct experiments with larger, high-quality datasets, as finding such data was difficult. Additionally, it is estimated that experimenting with larger models and larger datasets would require a significant amount of time, making these experiments infeasible. Therefore, while the ELO method demonstrated superior performance over FFT with datasets up to 200GB, further experiments with larger data sizes are necessary.

\section{Acknowledgment}
\label{sec:Acknowledgment}
We would like to thank the reviewer for their insightful feedback throughout the study. This research was supported by the LG CNS collaborative research project, “Domain Expansion via Adaptive Policy Acquisition in Multi-Agent Systems” and Institute of Information \& communications Technology Planning \& Evaluation (IITP) grant, funded by the Korea government (MSIT) (No.RS-2024-00456709, A Development of Self-Evolving Deepfake Detection Technology to Prevent the Socially Malicious Use of Generative AI). We have used GPUs from Artificial intelligence industrial convergence cluster development project funded by the Ministry of Science and ICT (MSIT, Korea) \& Gwangju Metropolitan City awarded to KyungTae Lim.

\bibliography{custom}

\appendix

\onecolumn
\appendix

\addtocontents{toc}{\protect\setcounter{tocdepth}{2}} 

\hypersetup{linkcolor=black}
\renewcommand{\contentsname}{} 
\vspace*{0.5cm} 
\setlength{\cftaftertoctitleskip}{6em} 

\begin{center}
    {\Huge \textbf{Appendix}} 
\end{center}

\setlength{\cftbeforesecskip}{1.5em} 
\setlength{\cftbeforesubsecskip}{1em} 

\tableofcontents
\clearpage

\section{Data Analysis}
\subsection{Details of the Benchmark Dataset}
The evaluation of the LLM was divided into quantitative and qualitative assessments \cite{zhou2023lima,choi-etal-2024-optimizing-language}. Quantitative evaluation involves scoring based on numerical metrics, which is done automatically. For instance, multiple-choice questions, such as true/false or four-option questions, were categorized under quantitative evaluation. The datasets used for this include MMLU, KoBEST, and MARC-ja. In contrast, qualitative evaluation was applied to tasks requiring the assessment of long-form answers, which were evaluated either by human judges or through automated evaluation using GPT. The datasets used for qualitative evaluation include MT-Bench, LogicKor, and MT-Bench(ja). Below is an introduction to the detailed evaluation datasets for each language.

\textbf{English}
\begin{itemize}
\item \textbf{MT-Bench:} MT-bench is a set of challenging 80 multi-turn open-ended questions for evaluating chat assistants . To automate the evaluation process, MT-bench prompts strong LLMs like GPT-4 to act as judges and assess the quality of the models' responses. The maximum score is 10 points.
\item \textbf{MMLU:} MMLU(Massive Multitask Language Understanding) is a benchmark that evaluates knowledge across 57 topics. In this paper, we used accuracy as the evaluation metric. 
\end{itemize}

\textbf{Korean}
\begin{itemize}
\item \textbf{LogicKor:} LogicKor is a multi-turn benchmark dataset designed to measure reasoning ability in various domains for Korean language models, using an LLM-as-a-judge approach. The dataset consists of a total of 42 multi-turn prompts across six categories: reasoning, mathematics, writing, coding, comprehension, and Korean language. LogicKor prompts strong LLMs like GPT-4 to act as judges and assess the quality of the models' responses. The maximum score is 10 points.
\item \textbf{KoBEST:} KoBEST is a Korean benchmark suite consists of 5 natural language understanding tasks that requires advanced knowledge in Korean. In this paper, we used F1-score as the evaluation metric. 
\end{itemize}

\textbf{Japanese}
\begin{itemize}
\item \textbf{MT-Bench(ja):} MT-Bench(ja) is a benchmark released by Stability-AI, created using the MT-Bench. MT-bench(ja) prompts strong LLMs like GPT-4 to act as judges and assess the quality of the models' responses. The maximum score is 10 points.
\item \textbf{MARC-ja:} MARC-ja is a dataset of the text classification task. This dataset is based on the Japanese portion of Multilingual Amazon Reviews Corpus. In this paper, we used accuracy\_norm as the evaluation metric. 
\end{itemize}

\subsection{Details of the Pretraining Dataset}
Table~\ref{tab:koen_pretraining_dataset_source},\ref{tab:jpen_pretraining_dataset_source} displays the sources of the datasets used for pretraining, along with the size of each dataset. In this paper, we express data size in GB rather than in tokens to avoid disparities in the number of samples across languages, which would hinder fair data utilization. The number of tokens per GB varies by language, ranging from approximately 1 billion to 1.3 billion. Thus, 10GB of data contains roughly 10 to 13 billion tokens.

\begin{table}[H]
\small
\centering
\begin{tabular}{lccc}
\toprule
        Language        & Source  & Content & Size(GB)\\
\midrule
\multirow{7}{*}{\textbf{Korean}}  &
\text{AI-Hub}\tablefootnote{https://www.aihub.or.kr} & News, Books & 19.15 \\ 
\cmidrule{2-4}
& \text{Modu-corpus}\tablefootnote{https://kli.korean.go.kr/corpus} & Paper, News & 19.20 \\ 
\cmidrule{2-4}
& \text{WIKI-ko}\tablefootnote{https://github.com/lovit/kowikitext} & Wikipedia & 1.17 \\ 
\cmidrule{2-4}
& \text{uonlp/CulturaX}\tablefootnote{https://huggingface.co/datasets/uonlp/CulturaX} & Web & 99.73 \\ 
\cmidrule{2-4}
& \text{cc100-ko}\tablefootnote{https://huggingface.co/datasets/lcw99/cc100-ko-only} & Web & 40.46 \\ 
\cmidrule{2-4}
& \textbf{Total} &  &  \textbf{179.71} \\ 
\midrule

\textbf{English} & 
\text{fineweb}\tablefootnote{https://huggingface.co/datasets/HuggingFaceFW/fineweb} & Web  & \textbf{19.98}\\ 
\midrule
\textbf{Total}        &     &  & \textbf{199.69} \\ 
\midrule
\end{tabular}
\caption{Korean Pretraining Dataset Source}
\label{tab:koen_pretraining_dataset_source}
\end{table}



\begin{table}[H]
\small
\centering
\begin{tabular}{lccc}
\toprule
Language        & Source  & Content& Size(GB) \\
\midrule
\textbf{Japanese} & 
\text{uonlp/CulturaX}\tablefootnote{https://huggingface.co/datasets/uonlp/CulturaX} & Web  & \textbf{9.00}\\ 
\midrule
\textbf{English} & 
\text{fineweb}\tablefootnote{https://huggingface.co/datasets/HuggingFaceFW/fineweb} & Web  & \textbf{1.00}\\ 
\midrule
\textbf{Total}        &     &  & \textbf{10.00} \\ 
\midrule
\end{tabular}
\caption{Japanese Pretraining Dataset Source}
\label{tab:jpen_pretraining_dataset_source}
\end{table}

\subsection{Details of the Instruction Tuning Dataset}
\label{sec:appendix-IT}
For a fair evaluation, we used the publicly available Instruction-Following dataset during the SFT (Supervised Fine-Tuning) phase, applying it uniformly across all models. The Tagengo dataset consists of over 70,000 prompt-response pairs in the ShareGPT format, covering 74 languages, formatted similarly to those used in Vicuna~\cite{chiang2023vicuna}. This dataset underwent human review and modification. We collected Korean, Japanese and English subsets from the Tagengo dataset, gathering 31K Instruction-Following pairs. Samples of the Korean data utilized can be found in Table~\ref{tab:sample_sftdata}.

\begin{table}[!ht]
\small
\centering
\resizebox{.95\textwidth}{!}{
\begin{tabular}{| p{0.95\textwidth} |}
\hline
\textbf{Sample of lightblue/tagengo-gpt4 data} \\
\hline
\textbf{[Korean Instruction]} \\
금융 전문가가 되어 주식이 무엇인지 설명하시오. \\
\quad(Please explain what stocks are as if you were a financial expert.)\\
\textbf{[OUTPUT]} \\
주식(stock)은 기업의 소유권을 나타내는 금융 자산입니다. 한 기업이 자본을 조달하기 위해 발행하는 주식을 구매함으로써, 구매자는 그 기업의 일부 소유권을 갖게 됩니다. 이는 구매자가 해당 기업의 주주가 되며, 기업의 이익과 손실, 성장과 위축에 따라 주식의 가치가 변동한다는 것을 의미합니다.\\
(A stock represents a financial asset that signifies ownership in a company. By purchasing shares issued by a company to raise capital, the buyer acquires partial ownership of that company. This means the buyer becomes a shareholder and the value of the stock fluctuates with the company's profits, losses, growth, and contraction.)\\
\\
주식은 크게 두 가지 유형으로 나뉩니다: 보통주와 우선주입니다.\\
(Stocks are primarily divided into two types: common stock and preferred stock.)\\
\\
\quad1. \textbf{보통주(Common Stock)}: 보통주는 주주에게 투표권을 부여하며, 기업의 이익에 따라 배당금을 받을 수 있는 권리를 줍니다. 그러나 회사가 파산할 경우, 채권자, 우선주 주주, 그리고 다른 의무가 충족된 후에야 보통주 주주들이 자산에서 몫을 받을 수 있습니다.\\
(1. \textbf{Common Stock}: Common stock grants shareholders voting rights and the right to receive dividends based on the company's profits. However, in the event of bankruptcy, common stockholders receive their share of assets only after creditors, preferred stockholders, and other obligations have been satisfied.)\\
\\
\quad2. \textbf{우선주(Preferred Stock)}: 우선주는 보통주보다 배당금 지급에 있어 우선권을 갖습니다. 즉, 회사가 배당금을 지급할 때 우선주 주주들이 먼저 배당을 받습니다. 그러나 대부분의 우선주는 투표권이 없어 회사의 중요한 결정에 참여할 수 없습니다. \\
(2. \textbf{Preferred Stock}: Preferred stock has priority over common stock in terms of dividend payments. This means preferred stockholders receive dividends first when the company pays out dividends. However, most preferred stocks do not grant voting rights, so holders cannot participate in major company decisions.)\\
\\
주식은 주식시장에서 거래됩니다. 이 시장은 크게 두 가지로 나뉘는데, 하나는 주식을 처음으로 공개하는 기업과 투자자 간의 거래가 이루어지는 \textbf{기업공개(IPO, Initial Public Offering)} 시장이고, 다른 하나는 이미 공개된 주식이 투자자들 사이에서 거래되는 \textbf{2차 시장}입니다.\\
(Stocks are traded on the stock market, which is divided into two main types: the \textbf{Initial Public Offering (IPO)} market, where shares are first issued and traded between the company and investors, and the \textbf{secondary market}, where already issued shares are traded among investors.) \\
\\
주식 투자는 기업의 성장에 참여하고 그 이익을 공유할 수 있는 기회를 제공하지만, 기업의 성과에 직접적으로 영향을 받기 때문에 위험도 내포하고 있습니다. 따라서 투자자는 주식 투자에 앞서 기업 분석, 시장 분석, 자신의 투자 목표와 위험 감수 능력을 면밀히 고려해야 합니다. \\
(Stock investing provides an opportunity to participate in a company's growth and share in its profits, but it also involves risk due to the direct impact of the company's performance. Therefore, investors should carefully consider company analysis, market analysis, their investment goals, and risk tolerance before investing in stocks.
)\\
\hline
\end{tabular}
}
\caption{\label{sft_data_sample}Example of the instruction data for ShareGPT data}
\label{tab:sample_sftdata}
\end{table}

\section{Experiment Environment}
To ensure reproducibility and comparability across studies, we conducted evaluations using publicly available benchmarking tools \cite{gao2021framework}.
\paragraph{GPUs Used.} 
We used eight NVIDIA H100 GPUs for the training and evaluation of the model.

\paragraph{Hyperparameters.} The hyperparameter settings used in this study can be found in Table~\ref{tab:hyperparameter}. All models were trained for 1 epoch during the PT stage and 10 epochs during the SFT stage.

\begin{table}[h] 
\centering
{\footnotesize
    \begin{tabular}{l | c } 
    \hline
{} & {value} \\ 
 \hline
 learning\_rate & 5.0e-06  \\
 Optimizer & AdamW\_bnb\_8bit \\ 
 lr\_scheduler & cosine \\
 Epoch for PT & 1  \\ 
 Epoch for SFT & 10  \\ 
 sequence\_len & 8192 \\
 Batch size for ELO PT  & 4  \\
 Batch size for ELO align  & 1  \\
 Batch size for sft  & 1  \\
 Random Seed & 42  \\ 
 \hline
  \end{tabular}
}
\caption{Applied hyperparameter settings.}   
\label{tab:hyperparameter}
\end{table}

\paragraph{Experiment Reproduction.}
We are making code used for testing available to allow for exact reproduction of the experiments conducted in this study. The qualitative responses generated by the models during the experiments can be downloaded from the supplementary materials.

\end{document}